\documentclass[lettersize,journal]{IEEEtran}
\usepackage{amsmath,amsfonts}
\usepackage{algorithmic}
\usepackage{algorithm}
\usepackage{array}
\usepackage[caption=false,font=normalsize,labelfont=sf,textfont=sf]{subfig}
\usepackage{textcomp}
\usepackage{stfloats}
\usepackage{url}
\usepackage{verbatim}
\usepackage{graphicx}
\usepackage[nocompress]{cite}
\usepackage{multirow}
\usepackage[normalem]{ulem} 
\usepackage{xcolor} 
\usepackage{booktabs}
\usepackage{amsmath}  
\usepackage{amssymb} 
\usepackage{bm}      
\usepackage{dsfont}   
\begin{document}

% \title{A Sample Article Using IEEEtran.cls\\ for IEEE Journals and Transactions}
\title{UniForensics: Face Forgery Detection via\\ General Facial Representation}

\author{Ziyuan~Fang$^*$,~Hanqing~Zhao$^*$,~Tianyi~Wei,~Wenbo~Zhou$^\dagger$,~Ming~Wan,~Zhanyi~Wang,~Weiming~Zhang,Nenghai~Yu
        % <-this % stops a space
\IEEEcompsocitemizethanks{
\IEEEcompsocthanksitem {This work was supported in part by the Natural Science Foundation of China under Grant  62121002,62372423,U2336206,U20B2047, and 62072421, Key Research and Development program of Anhui Province under Grant 2022k07020008.}
\IEEEcompsocthanksitem Ziyuan Fang, Hanqing Zhao, Tianyi Wei, Wenbo Zhou, Weiming Zhang, Nenghai Yu are with University of Science and Technology of China, Hefei, Anhui 230026, China. E-mail: \{umbreller@mail., zhq2015@mail., bestwty@mail., welbeckz@, zhangwm@, ynh@\}ustc.edu.cn
\IEEEcompsocthanksitem Ming Wan, Zhanyi Wang are with Qianxin Technology Group Co. Ltd, Beijing, China. Email: \{wanming,wangzhanyi\}@qianxin.com
\IEEEcompsocthanksitem $^*$ Ziyuan Fang and Hanqing Zhao are co-first authors, $\dagger$ Wenbo Zhou is the corresponding author.
}
}

\markboth{Journal of \LaTeX\ Class Files,~Vol.~14, No.~8, August~2021}%
{Shell \MakeLowercase{\textit{et al.}}: A Sample Article Using IEEEtran.cls for IEEE Journals}

\maketitle

\begin{abstract}
The rise of deepfakes has significantly heightened concerns for privacy and the authenticity of digital media, bringing widespread attention to face forgery detection.
Previous deepfake detection methods mostly depend on low-level textural features vulnerable to perturbations and fall short of detecting unseen forgery methods. In contrast, high-level semantic features are less susceptible to perturbations and not limited to forgery-specific artifacts, thus having stronger generalization. Motivated by this, we propose a detection method that utilizes high-level semantic features of faces to identify inconsistencies in temporal domain.
We introduce UniForensics, a novel deepfake detection framework that leverages a transformer-based video classification network, initialized with a meta-functional face encoder for enriched facial representation. In this way, we can take advantage of both the powerful spatio-temporal model and the high-level semantic information of faces. Furthermore, to leverage easily accessible real face data and
guide the model in focusing on spatio-temporal features, we design a Dynamic Video Self-Blending (DVSB) method to efficiently generate training samples with diverse spatio-temporal forgery traces using real facial videos. Based on this, we advance our framework with a two-stage training approach: The first stage employs a novel self-supervised contrastive learning, where we encourage the network to focus on forgery traces by impelling videos generated by the same forgery process to have similar representations.
On the basis of the representation learned in the first stage, the second stage involves fine-tuning on face forgery detection dataset to build a deepfake detector. 
Extensive experiments validates that UniForensics outperforms existing face forgery methods in generalization ability and robustness. In particular, our method achieves 95.3\% and 77.2\% cross dataset AUC on the challenging Celeb-DFv2 and DFDC respectively.

\end{abstract}


\section{Introduction}
\IEEEPARstart{T}{he} advent of face manipulation technologies has led to the creation of deepfakes that are virtually indistinguishable from authentic images, posing profound risks to individual privacy and the integrity of public discourse. As a result, the development of sophisticated detection methods has become a critical necessity. The current challenge of face forgery detection is twofold: accurately identifying forgeries that closely mimic genuine content and, more importantly, ensuring that methods can generalize to detect unseen manipulation techniques. Consequently, there is an urgent need to advance the generalization capabilities of deepfake detection methods so that we can maintain trust in digital media.

Early face forgery detection methods\cite{qian2020thinking,li2020face,liu2021spatial} mainly relied on low-level textural features. However, these features are susceptible to digital corruptions such as video compression and becoming more and more difficult to distinguish with the advancement of forgery methods. 
In face manipulation, compared to generating lifelike facial images, it is more challenging to decouple identity attributes from expression attributes and to recombine features of different individuals, due to the lack of ground truth supervision. This leads to the confusion between the high-level semantic features of the source face and the target face during the face forgery process. Therefore, it is crucial to leverage high-level semantic features to improve the generalization ability of face forgery detection. However, there is a challenge in such approach: with the rapid development of forgery methods, fake faces are highly similar to real ones in visual perception, so it is not easy to distinguish authenticity from forgeries simply using high-level semantic features within a single frame. To address this, we target at the temporal inconsistency of high-level semantic features in deepfake videos.

Because most of the current forgery methods are frame-based while ignoring the temporal consistency, video-level detection holds significant potential in enhancing generalization capabilities\cite{zheng2021exploring,guan2022delving,zhao2023istvt,xu2023tall,wang2023altfreezing}. Compared with image-level deepfake detection, video-level detection can not only use spatial-related artifacts,
but also use artifacts in temporal domain, such as identity inconsistency and expression discontinuity. These high-level semantic artifacts are caused by the deficiencies in the disentangling capabilities of the face forgery model (For instance, the proportions of the mixture of source and target ID information differ across frames.) and are less susceptible to the change of visual quality, thus possessing better generalization capabilities. Based on this assumption, some previous methods have tried to use specific semantic feature for video-level forgery detection, such as lip movement features\cite{haliassos2021lips} and 3D structural features\cite{cozzolino2021reveal}.

Different from these methods, we introduce a more general and robust face representation from a pre-trained meta-functional face encoder. This high-level semantic feature contains the identity, geometric structure, expression and richer attributes of the face, contributing to the model's capacity to discern inconsistency in identity features and motion patterns across video frames, thus aiding in the accurate identification of falsified videos. It is notable we are the first to discover that the ability of a pre-trained face encoder can be effectively transferred to the task of deepfake detection.

In this paper, we introduce UniForensics, a novel framework that uncover the potential of general facial representation for face forgery detection. We use a transformer-based video classification network with separable temporal and spatial related parameters and initialized its spatial module with a pre-trained face encoder. By doing so, we combined a general and robust facial representation with a powerful spatio-temporal modeling framework, allowing the model to benefit from both aspects simultaneously.

To further guide the model in focusing on spatio-temporal features and utilizing a large amount of easily accessible real face data, we design a method to efficiently generate training samples with diverse spatio-temporal forgery traces using real facial videos. We draw inspiration from the technique of self-blended image\cite{shiohara2022detecting} which has shown significant promise for image-level face forgery detection. This approach synthesizes training samples by blending real facial images to imitate the general deepfake generation process. However, when extended to video-level deepfake detection, if we use independent transform argument for each frame, the generated video will exhibit significant jitter and the model is prone to overfit. On the contrary, if the same transform argument is applied for each frame, the model will tend to detect monotonous spatial features, thereby weakening its sensitivity to temporal inconsistencies. To address this limitation, we introduce a novel dynamic video self-blending method that enables us to control the temporal discrepancy. This technique generates synthetic fake samples with a broader spectrum of temporal artifacts, thereby enhancing the generalization capacity of our deepfake detection model. We term this technique as Dynamic Video Self-Blending (DVSB).

Using the fake samples generated by DVSB, we can expose the model to a wide range of forgery traces. However, we find that training solely on a binary classification task is not the optimal approach, especially when using synthesized fakes that cover a diverse array of forgery traces. Therefore, we propose a two-stage training framework to exploit the potentialities of DVSB in order to refine the model's detection capabilities. During the first stage of training, we employ a novel self-supervised contrastive learning approach using generated fake samples. Specifically, we ensure that videos produced by the same forgery process have similar representations. Our purpose is to make the network focus on a more diverse range of forgery traces, thereby achieving a better representation for deepfake detection. Subsequently, on the basis of this well-learned representation, we apply a fine-tuning stage on a dedicated face forgery detection dataset to refine the model's detection capabilities.

Extensive experiments substantiate that our UniForensics achieves state-of-the-art performances in the field, exhibiting superior generalization across unseen manipulation methods and datasets. It demonstrates remarkable accuracy in detecting forgeries, even in videos of lower quality, and displays a strong robustness to a variety of digital corruptions. Through comprehensive ablation studies, the effectiveness and strategic acumen of UniForensics has been affirmed.

Our contributions can be summarized as follows:

\begin{itemize}
  \item We proposed UniForensics, a novel face forgery detection framework that integrates a transformer-based video classification network with the enriched facial representation of a meta-functional face encoder. Thus, our UniForensics can benefit from both the powerful spatio-temporal model and the high-level semantic information of faces, achieving better generalization ability and robustness.
  \item We designed a dynamic video self-blending (DVSB) method to synthesize fake samples with diverse and authentic spatio-temporal artifacts. It leverages readily available real face data to guide the model in focusing on spatio-temporal features.
  \item We introduced a two-stage training strategy for video deepfake detection: The first stage employs a novel self-supervised contrastive learning where the model is trained to identify pairs of samples that have undergone the same forgery process. The second stage involves fine-tuning on face forgery detection dataset, aiming to enhance the model’s detection accuracy.
  \item We conducted extensive experiments on various Deepfake datasets including FaceForensics++ \cite{rossler2019faceforensics++}, FaceShifter\cite{li2020advancing}, Celeb-DFv2 \cite{li2020celeb} and DFDC\cite{dolhansky2020deepfake} datasets. The results demonstrate the effectiveness of our method.
\end{itemize}

\section{Related works}

Although many face forgery detection methods \cite{frank2020leveraging,zhao2021multi,nirkin2021deepfake,sun2021domain,ni2022core,gu2022exploiting} have been introduced in the past few years, their generalization to unseen forgery type still remains unsatisfactory. To address this, researchers have proposed a variety of methods \cite{chen2022ost,chen2022self,cao2022end,yang2023masked,dong2023implicit} to enhance the generalization of deepfake detection model. 
After reviewing masses of existing methods, we found that there are three important categories of methods that promote the performance of deepfake detection algorithms on unseen forgery type: model pre-training, data synthesis and augmentations, and spatio-temporal learning.

\subsection{Model Pre-training}

Pre-training is an important approach in deep learning that can provide representation with rich information and high robustness for downstream tasks. Based on the rich semantic information embedded in pre-trained features, researchers can develop detection models with better performance. More importantly, the powerful prior knowledge embedded in pre-trained models can mitigate overfitting to low-level forgery artifacts to a certain extent, thus enhancing the model's generalization ability to unseen forgery types and robustness to visual perturbations.

Previous works have proved that pre-training on general image classification task can significantly enhance the performance of deepfake detection models. For example, SBI \cite{shiohara2022detecting}, OST \cite{chen2022ost} and TALL \cite{xu2023tall} all adopted a backbone pre-trained on ImageNet. This can be attributed to that it provides the model with better initialization for image classification. Researchers have also tried to leverage models that pretrained on facial images. AUNet\cite{bai2023aunet} used a pretrained face parsing model to guide the model to exploit relation between different action units of human face. \cite{shi2023real} Shi et al. proposed to pretrain a model on real face datasets by masked image modeling (MIM) and detect deepfakes through reconstruction discrepancy.

Taking it a step further, Haliassos et al. proposed LipForensics \cite{haliassos2021lips} that specifically pretrained a spatio-temporal network to perform visual speech recognition (lipreading) and finetuned it for deepfake detection. As the successor, RealForensics \cite{haliassos2022leveraging} used temporally dense video representations pretrained in a self-supervised crossmodal manner to guide the detection model. These two methods provide the model with high-level semantic information through cross-domain pre-training, thus enhancing the generalization ability of detection model. However, the potential of general facial representation on deepfake detection has not been fully explored yet. In this work, we introduce a more general face representation from a pre-trained face encoder, equiping our detection model with high-level facial features. In addition, we design a self-supervised pretraining scheme to guide the model to learn more general and diverse forgery traces.

\subsection{Data Synthesis and Augmentations}

Due to the rapid advancement of facial forgery techniques, the methods collected in existing datasets are quickly being surpassed by newer, more realistic ones. Consequently, models built upon these datasets often experience significant performance drop when applied across unseen datasets. 

Resent methods \cite{li2020face,zhao2021learning,shiohara2022detecting,chen2022self} have proved that synthesized fake samples with more general and hardly recognizable artifacts can encourage classifiers to learn generic and robust representations and therefore prevent detection model from overfitting to manipulation specific artifacts. Face X-ray \cite{li2020face}, I2G \cite{zhao2021learning} and SLADD \cite{chen2022self} propose to synthesize fake images by blending two images to simulate the blending boundary artifacts that shared by most forgery methods. Building upon these methods, SBI \cite{shiohara2022detecting} further propose to blend source and target images generated from one real image, mitigating the negative impact of different face id. Recently, SeeABLE \cite{larue2023seeable} proposed to generate fine-grained local image anomalies between different patches and the rest of the image, and train a one head multi-objective regressor to regress the localization error of the generated soft discrepancies on disrupted faces. In AltFreezing \cite{wang2023altfreezing}, Wang et al. noticed that there is a lack of effective video-level data synthesis method in face forgery detection and designed a video-level data augmentation method with temporal dropout, repeat and clip level blending. Despite their good performance, there is still a lack of effctive method that simulates the temporal characteristics introduced by facial forgery techniques. In this work, we propose dynamic video self-blending to enrich the temporal forgery artifacts of synthesized fake videos. We also use dynamic video self-blending for self-supervised pretraining, aiming to help the model focus more on the generic manipulation artifacts.

\subsection{Spatio-temporal Learning}

To improve the generalization of deepfake detectors, researchers also proposed to capture the temporal incoherence of fake videos as the generic clues. 

Recent emerging studies have shown that spatio-temporal learning has become an important method for training deepfake detectors. FTCN \cite{zheng2021exploring} directly trained a fully temporal 3D ConvNets with an attached temporal transformer. Guan et al. proposed LTTD \cite{guan2022delving} that focuses on the valuable temporal information within local sequences using a well-designed transformer-based model. In ISTVT \cite{zhao2023istvt}, Zhao et al. used a novel decomposed spatial-temporal self-attention and a self-subtract mechanism to capture spatial artifacts and temporal inconsistency. TALL \cite{xu2023tall} transformed a video clip into a pre-defined layout to realize the preservation of spatial and temporal dependencies. AltFreezing \cite{wang2023altfreezing} proposed to alternately freeze spatial-related and temporal-related weights during the training process so that the model can learn spatial and temporal features. However, using spatio-temporal learning often makes it difficult to find suitable pretrained models, forcing us to train the networks from scratch. In this work, we used a transformer-based video classification network with separable temporal and spatial related parameters and initialized its spatial module with a pre-trained face encoder. 

\section{Method}

\begin{figure*}[t]
  \centering
  \includegraphics[width=0.95\linewidth]{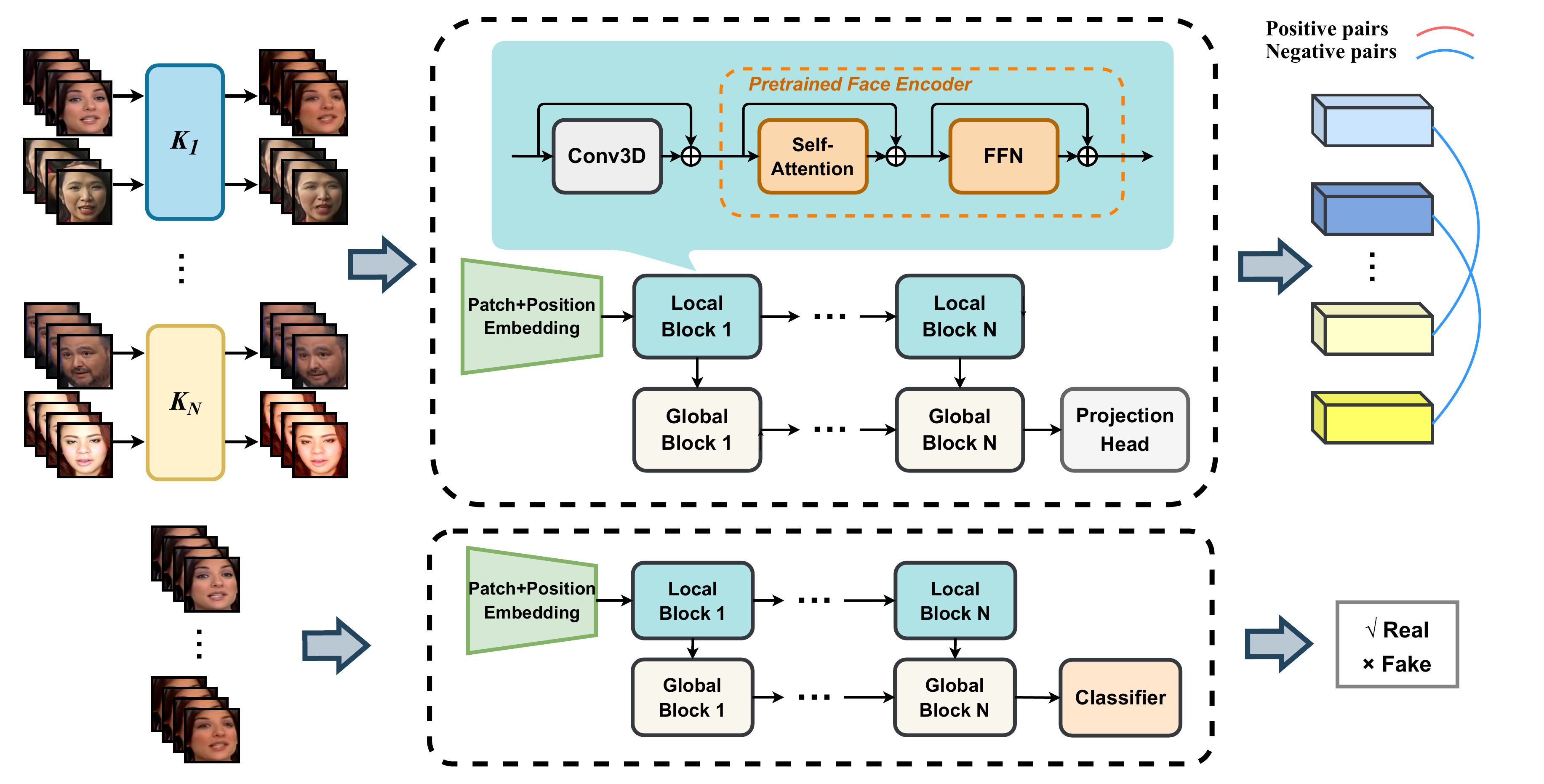}

   \caption{Overview of our proposed method. We use a pretrained meta-functional face encoder to initialize a transformer-based video classification network as detection backbone and advance it with our two-stage training strategy. The first stage employs self-supervised contrastive learning on fake samples generated from real face data. The second stage involves fine-tuning on deepfake detection dataset.}
   \label{fig:main}
\end{figure*}

\subsection{Overview}

Previous face forgery detection methods typically relied on identifying low-level visual artifacts, making it difficult to resist image compression and advanced forgery methods. For improving generalization, we utilize a pretrained meta-functional face encoder that extracts semantic information of face images in detail to initialize a transformer-based video classification network as detection backbone. Incorporated with the high-level semantic feature, the model can learn to discern inconsistency in identity and motion patterns across video frames, leading to better generalization performance. Furthermore, to leverage easily accessible real face data and guide the model in focusing on spatio-temporal features, we design a method to efficiently generate training samples with diverse spatio-temporal forgery traces using real facial videos. However, we find that simply training on binary classification task is not the optimal choice especially when using synthesized fakes that cover a wide range of forgery traces. So we advance our framework with a two-stage training scheme(as depicted in Fig. \ref{fig:main}): The first stage employs self-supervised contrastive learning on a real facial video dataset by impelling videos generated by the same forgery process to have similar representations. Based on the well-learned representation from the first stage, the second stage involves fine-tuning on a face forgery detection dataset, aiming to enhance the model's detection accuracy.

\subsection{Model Initialization}

Early face forgery detection methods typically relied on low-level textural features. However, these features are susceptible to digital corruptions such as video compression and increasingly difficult to distinguish with the rapid advancement of forgery techniques. Therefore, it is crucial to leverage high-level semantic features to enhance the generalization ability of deepfake detection. On account of this, we introduce a more general and robust face representation from a pre-trained meta-functional face encoder, FaRL\cite{zheng2022general}. FaRL is a ViT-based general facial encoder trained through image-text contrastive learning and masked image modeling. It is pre-trained on LAION-FACE, a dataset containing a large amount of face image-text pairs, using the two aforementioned strategies. We aim to train a powerful deepfake detection model based on FaRL's general and robust representations.

However, as face forgery becomes increasingly realistic, it is not easy to distinguish authenticity from forgeries simply using high-level semantic features within a single frame. Therefore, we target at the temporal inconsistency of high-level semantic features in deepfake videos. We refer to UniFormerV2\cite{li2022uniformerv2}, which 
is a transformer-based video classification network with separable temporal and spatial related parameters and its spatial module can be initialized with a pre-trained ViT. To be specific, we first use 3D convolution to project the input video as spatio-temporal tokens. Then replace the convolution in the Transformer block with 3D convolution while retaining the original spatial self-attention and FFN layers, resulting in a local block primarily responsible for processing spatial information. To further achieve global spatio-temporal information modeling, we construct global block in a similar way, while using the output of the local block as keys and values and compute global spatio-temporal attention instead of spatial one. In this way, the self-attention and FFN layers in the local block can utilize the pre-trained weights from the ViT model. We initialize the UniFormerV2 backbone using the pre-trained ViT model of FaRL.

The face encoder of FaRL can extract rich high-level semantic features from facial images such as geometric structure, identity and expression. Different from the low-level texture features that many previous face forgery detection methods rely on, these semantic features are more robust to video distortions. Once these features are exploited by UniFormerV2, they will lead to strong detection performance, as UniFormerV2's powerful spatio-temporal modeling capabilities can discern anomalies in the temporal sequence of these features. The pretrained semantic features will guide the model to discriminate more based on high-level semantic informations, thereby achieving better robustness and transferability. Given that the majority of current forgery methods are frame-based,
the face-swapping process can easily cause inconsistency in semantic information between frames and discontinuity in movement. Our design aim to leverage UniFormerV2's temporal modeling capability to detect these temporal anomalies and build a video-level deepfake detector upon the image-level features of FaRL.

\begin{figure*}[t]
  \centering
  \includegraphics[width=0.95\linewidth]{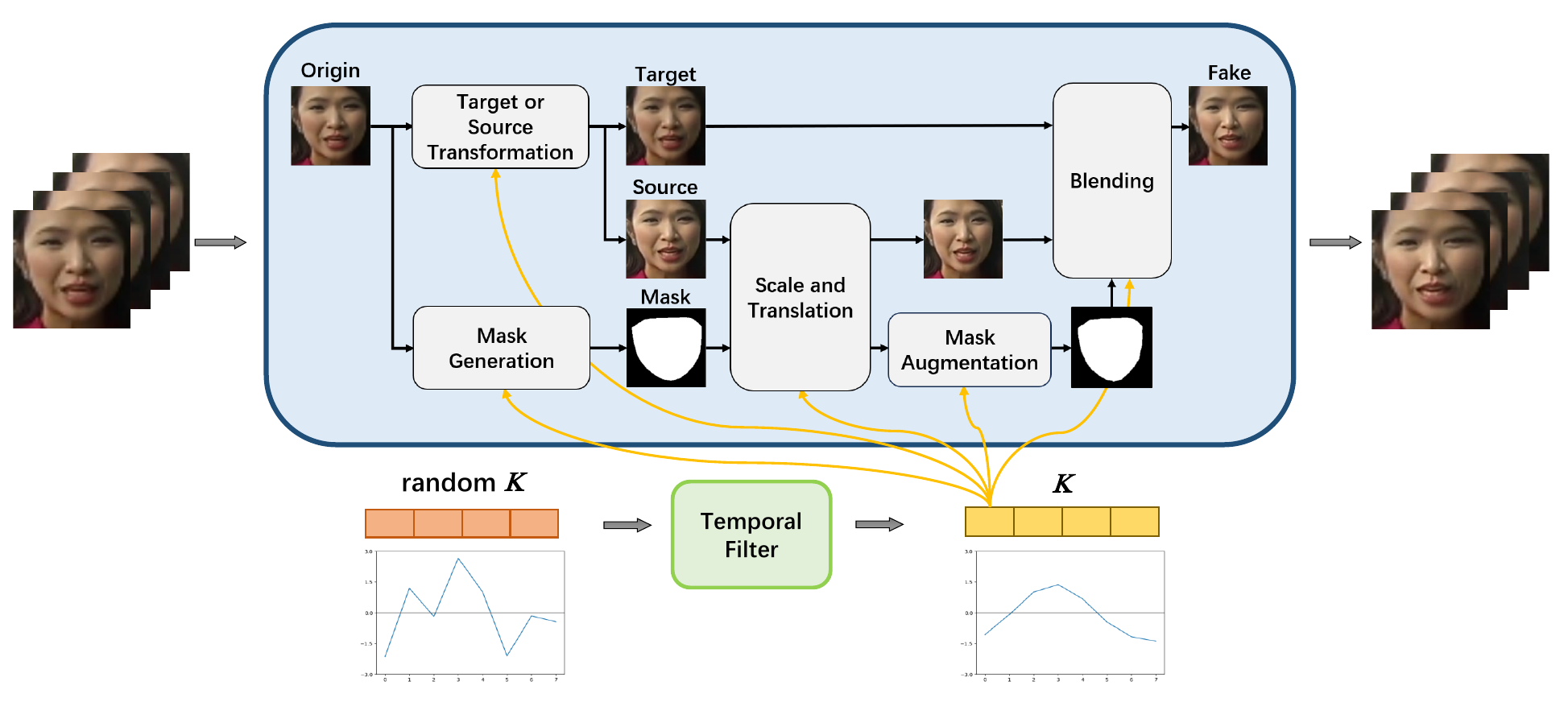}

   \caption{\textbf{Dynamic Video Self-Blending: }$K$ is a group of parameters controls realization of transforms to the target image and blending policies. We firstly generate a random sequence of $K$ and then use a temporal filter to make the transformation $K$ of each frame smooth in time. Consequently, the synthesized fake video has non-uniform temporal artifacts.}
   \label{fig:DVS}
\end{figure*}

\subsection{Dynamic Video Self-Blending (DVSB)}

\textbf{Preliminaries} SBI \cite{shiohara2022detecting} is an effective data augmentation method using image self-blending to synthesize fakes. It first duplicated the image into two instances, one designated as the source and the other as the target. Then randomly choose the target or source image and conduct transformations on the color distribution and quality of it to generate the difference between target and source. Meanwhile, extract facial keypoints to obtain a mask of the facial area. After that, perform same scale and translation on the source image and facial mask to imitate the dislocation of blending and perform augmentations on the mask. Finally, blend the source with the target to produce a forgery-like image.

It has been proved that carefully crafted synthesized fake samples are beneficial to classifier's generalization ability. But when it comes to video level, existing methodologies are insufficient as they either solely engage in image-level processing \cite{li2020face,shiohara2022detecting}, overlooking spatial features, or rely on simplistic methods such as frame interpolation and dropping \cite{wang2023altfreezing}. So, what kind of video-level data synthesis and augmentation is appropriate in deepfake detection? Let us begin by analyzing single-frame operations.

For each frame of a video segment, our operations are listed as follows: 

1. Color distribution adjustment controlled by parameter $K_c^{(t)}$ (where $K_c$ includes offsets for RGB, HSV, contrast and brightness, and t represents the frame number in the video)

2. Image quality adjustment controlled by parameter $K_q^{(t)}$ (where $K_q$ is the downscale rate or convolution kernel of the sharpening filter)

3. Imitation of blending dislocation controlled by parameter $K_d^{(t)}$ (which are the scale and translation proportions in x and y coordinates).

4. Mask generation via taking the convex hull of face landmarks controlled by parameter $K_g^{(t)}$ (which is the landmarks choosing scheme).

5. Mask augmentations via geometric elastic deformation adopted in \cite{zhao2021learning}, controlled by parameter $K_a^{(t)}$ (which are the offsets for each pixel's x and y coordinates).

6. Blending strategy, controlled by parameter $K_b^{(t)}$ (which are the number of iterations for applying Gaussian blur to the mask and the parameters of the Gaussian kernel).

The forgery trace for each video can be represented as:

\begin{equation}
\label{eq:video forgery}
  K^{(t)} = K_b^{(t)} \circ K_a^{(t)}\circ K_g^{(t)} \circ K_d^{(t)} \circ K_q^{(t)} \circ K_c^{(t)}  
\end{equation}

Through this series of transformations, we can convert each frame of the video into a self-blended image with traces of forgery. However, a question then arises: how to determine the relationships between operations on frames within a video segment, which will dictate the temporal characteristics of the generated samples? It is straightforward to think of two naive approaches: completely static or entirely random over time, but both exhibited significant shortcomings in practice, which we will analyze in detail later.

If we apply same transformation $K^{(t)}$ to every frame, the forgery characteristics would be static over time, and can be considered a direct current signal. The power spectrum of such a signal will be a delta function located at zero frequency because the signal contains no varying frequency components. 

In parallel, if the transformation $K^{(t)}$ for each frame is randomly sampled, the temporal forgery characteristics varies at a high frequency. The power spectrum is a constant value, indicating that the signal has the same energy distribution across all frequencies.

As depicted in Fig. \ref{fig:DVS}, to ensure the diversity of temporal characteristics, we proposed to do Gaussian filter on the white-noise-like $K^{(t)}$ before applying it to a video. The forgery parameters with temporal features are generated through the following filter process:

\begin{equation}
\label{eq:dvsb}
    K_f=IDCT(F(DCT(K))) 
\end{equation}

where $K = \{K^{(t)}\}, t \in \{1,2,...,T\}$, and $T$ is the length of forgery clip. $DCT$ and $IDCT$ are discrete cosine transform and inverse discrete cosine transform respectively. They are responsible for transforming K into the frequency domain and restoring it back. $F$ is the Gaussian filter with random mean and variance. Throughout the filtering process, we maintain the total power of the signal unchanged. Specifically, the Gaussian filter $F$ we employ is defined as follows:

\begin{equation}
\label{eq:gs filter}
    g(x) = \frac{1}{\sigma \sqrt{2\pi}} e^{-\frac{(x - \mu)^2}{2\sigma^2}}
\end{equation}
\begin{equation}
    \{D^{(t)}\} = DCT(K), t \in \{1,2,...,T\}
\end{equation}
\begin{equation}
    P = \sum_{t=1}^{T} \vert D^{(t)}\vert ^2
\end{equation}
\begin{equation}
    \hat{D}^{(t)}_f = g(t)\cdot D^{(t)}, t \in \{1,2,...,T\}
\end{equation}
\begin{equation}
    C_{scale} = \sqrt{\frac{P}{\sum_{t=1}^{T} \vert \hat{D}^{(t)}_f\vert ^2}}
\end{equation}
\begin{equation}
   D^{(t)}_f = C_{scale}\cdot \hat{D}^{(t)}_f, t \in \{1,2,...,T\}
\end{equation}

where $P$ is the power of the original signal $D^{(t)}$, $C_{scale}$ is the coefficient used to maintain the signal power unchanged before and after filtering. $\mu$ and $\sigma$ are the mean and variance of the Gaussian distribution. They are sampled independently from two uniform distributions respectively: $\mu\sim U[-3,3]$ and $\sigma\sim U[1,5]$.

We applied our dynamic video self-blending during our two-stage training scheme but the configurations of $K$ differed between stages. It is important to note that all generated samples used in our method have undergone filtering. For the sake of convenience in notation, we have omitted the subscript $f$ of $K_f$.

In addition, inspired by \cite{jiang2020deeperforensics}, we introduce a masked adaptive instance normalization (MAdaIN) to narrow the gap in color distribution between target and source image, thus making the samples more challenging to distinguish. Before blending the source with the target, we use MAdaIN to adaptively compute the affine parameters from the face area of target image:

\begin{equation}
    \left\{
    \begin{array}{l}
   s = m\cdot I_s \\
   t = m\cdot I_t
   \end{array}
    \right.
\end{equation}
\begin{equation}
   s' = \sigma(t)\frac{s-\mu(s)}{\sigma(s)}+\mu(t)
\end{equation}

where $I_s$ and $I_t$ are the source and target image before blending, $m$ is the facial mask generated by $K_g^{(t)}$ and $K_a^{(t)}$. $\sigma(\cdot)$ and $\mu(\cdot)$ stand for the calculation of mean and variance respectively. 

After the operation of MAdaIN, we blend the source with the target using $K_b^{(t)}$. It is worth noting that in practice we apply the MAdaIN to the samples with a specified probability.

\subsection{Two-Stage Training Scheme}

With the fake samples generated by DVSB, we can expose the model to a wide range of forgery traces. However, we find that training solely on a binary classification task is not the optimal approach, especially when using synthesized fakes that cover a diverse array of forgery traces. The fake samples generated by DVSB are able to provide stronger supervision for model training compared to binary classification. Therefore, we propose a two-stage training framework to leverage the potential of DVSB in order to refine the model’s detection capabilities. The first stage employs self-supervised contrastive learning on a real facial video dataset by encouraging videos generated by the same forgery process to have similar representations. Building on the well-learned representations from the first stage, the second stage involves fine-tuning on a face forgery detection dataset to enhance the model’s detection accuracy.

\subsubsection{Self-supervised Contrastive Pretrain (Forgery Process Identification)}

In the first stage of training, we conduct a self-supervised contrastive learning adapted from SimCLR\cite{chen2020simple}. Different from SimCLR, we do not treat the augmented pairs of one data example as positive. We use our proposed dynamic video self-blending to generate positive pairs with forgery traces of same category and treat the other generated examples within a minibatch as negative examples. During the contrastive learning stage, the network is supposed to extract generic temporal and spatial domain-related forgery traces from videos and represent them accordingly. We term this forgery aware pretraining as forgery process identification. If the output features of the model are able to categorize various forgery methods, we believe that the information it contains is also sufficient to identify whether or not a video has been forged. 

Following SimCLR, we introduce a learnable projection head between the representation and the contrastive loss We use the NT-Xent loss function for a positive pair of examples $(i, j)$, which is defined as:

\begin{equation}
\label{eq:NT-Xent}
\ell_{i,j} = -\log \frac{\exp(\mathrm{sim}(\bm z_i, \bm z_j)/\tau)}{\sum_{k=1}^{2N} \mathds{1}_{[k \neq i]}\exp(\mathrm{sim}(\bm z_i, \bm z_k)/\tau)}~,
\end{equation}

where $\mathds{1}_{[k \neq i]} \in \{ 0,  1\}$ is an indicator function evaluating to $1$ iff $k \neq i$ and $\tau$ denotes a temperature parameter and $N$ is the size of a minibatch.  After training is completed, we throw away the projection head and use the representation for deepfake detection task.

\subsubsection{Positive Pairs Generation} In the self-supervised pretraining stage, our objective is to encourage the network to learn and produce generic spatio-temporal forgery features. Hence, the crux of the issue lies in the design of positive samples. In our proposed dynamic video self-blending, color distribution adjustment $K_c$ and image quality adjustment $K_q$ is intended to generate the differences between target and source image. However, in the actual forging process, the differences in color distribution and image quality between the source and target are primarily attributed to the inherent characteristics of the images rather than the choice of forgery method. Conversely, the boundary features arising from image blending are strongly correlated with the choice of forgery method, which can be used as a generic artifact for detection. Therefore, we chose all random parameters associated with the blending process to generate positive pairs. Specifically, we utilize a same set of parameters $\{K_d, K_g, K_a, K_b\}$ to generate positive sample pairs, while allowing parameters $K_c$ and $K_q$ to vary randomly. To generate a sufficiently diverse set of blending features, we also expanded the range of values for parameter $K_g$. We divided the face landmarks into the left eye, right eye, nose and mouth. Then we used these subsets of landmarks and their combinations to generate partial face masks, thus simulating partial face blending.

During training, we randomly sample from the parameter space $\{K_d, K_g, K_a, K_b\}$ to obtain forgery transformations. Each transformation is applied to two real facial videos to generate positive pairs. We treat the other generated examples within a minibatch as negative examples. 
Similar to SimCLR, we add a projection head that maps representations to the space where contrastive loss is applied to our backbone. During training, the model is supposed to extract forgery-specific features from each video and make classification. As our positive and negative samples are both generated by sampling parameters from a random space, the model encounters a diverse range of forged features during the training process. This ensures that the features generated by the model can capture traces of various forgery methods, thereby enhancing its generalization capability.

\subsubsection{Supervised Finetuning on Face Forgeries}

After the pretraining in the first stage, the model has already established a well-trained representation of forgery traces. Then in the second stage, we replace the projection head with a MLP classifier head for deepfake detection. We conduct supervised finetuning on face forgeries to enhance our model's detection accuracy. We also also utilized forged samples generated through dynamic video self-blending as additional training data, in order to mitigate the model's overfitting to method-specific artifacts. In this stage, we simply sample transformation $K$ to synthesize fakes. That is, we randomly sample from the whole parameter space $\{K_c, K_q, K_d, K_g, K_a, K_b\}$ of $K$, since we don't need positive pairs with same forgery transformations anymore in this stage.

\section{Experiments}

\subsection{Experimental Settings}

\subsubsection{Datasets} For our self-supervised pretraining, we used 5994 video clips with different ids in VoxCeleb2\cite{chung2018voxceleb2}. We adopt the widely used deepfake detection benchmark FaceForensics++ \cite{rossler2019faceforensics++} (FF++) for supervised finetuning, following the convention. It contains 1,000 original videos and 4,000 fake videos forged by four manipulation methods: Deepfakes\cite{deepfake} (DF), Face2Face\cite{thies2016face2face} (F2F), FaceSwap\cite{faceswap} (FS), and NeuralTextures\cite{thies2019deferred} (NT). Besides, FF++ contains multiple video qualities, e.g. high quality (HQ), low quality (LQ) and RAW. We use the raw version for our training. For our cross-dataset evaluation, we use 3 recent challenging deepfake datasets as testing sets. (1) Celeb-DFv2 \cite{li2020celeb} (CDF) applies a more advanced deepfake technique to celebrity videos downloaded from YouTube. (2) DFDC\cite{dolhansky2020deepfake} is a large-scale benchmark developed for Deepfake Detection Challenge. This dataset includes 124k videos from 3,426 paid actors. The existing deepfake detection methods do not perform very well on DFDC due to their sophisticated deepfake techniques and shooting scenarios. (3) FaceShifter\cite{li2020advancing} is a face forgery dataset obtained by applying the FaceShifter manipulation method to the original video of FF++. We use the test videos, according to the FF++ split. 

\subsubsection{Implementation Details} For each video frame, face crops are detected by using RetinaFace\cite{deng2019retinaface} and landmarks are detected by the public toolbox Dlib\cite{king2009dlib}. All face crops are resized to 224 × 224 before being input. AdamW optimization is used with a learning rate of 1.0e-5 and batch size of 32, using a cosine decay learning rate scheduler. In the second stage of training, we set the learning rate of the classifier as 3.0e-5 and linearly decay it to 1.0e-5. The learning rate remains consistent with the backbone in subsequent epochs. We adopt the AUC (Area Under Receiver Operating Characteristic Curve) as the evaluation metrics for extensive experiments. We use FaRL's ViT pre-trained on LAION Face 20M and UniFormerV2-B/16 to initialize our model. 

That is, we insert the FaRL pretrained weights into the local block of UniFormerV2-B/16 and randomly initialize the remaining layers. In first stage of training, we augmented the model with a projection head, facilitating contrastive learning for the forgery process identification in a 256-dimensional feature space. In the second stage, we removed the projection head and equipped the model with a binary classification head, thereby adapting it for the task of deepfake detection. We apply the MAdaIN in DVSB with a probability of $25\%$.

\begin{table}
\caption{Generalization to unseen manipulations methods on FF++ dataset. We report video-level ROC-AUC (\%) on detecting each face forgery method excluded from training set.}
\centering
\begin{tabular}{lccccc}
\hline 
\multirow{2}*{Method} & \multicolumn{4}{c}{Trained on remaining three} & \multirow{2}*{Avg}\\ 
\cmidrule(lr){2-5} & DF & FS & F2F & NT \\
\hline Xception \cite{rossler2019faceforensics++} & 93.9 & 51.2 & 86.8 & 79.7 & 77.9 \\
CNN-aug \cite{wang2020cnn} & 87.5 & 56.3 & 80.1 & 67.8 & 72.9 \\
PatchForensics \cite{chai2020makes} & 94.0 & 60.5 & 87.3 & 84.8 & 81.7 \\
Face X-ray \cite{li2020face} & 99.5 & 93.2 & 94.5 & 92.5 & 94.9 \\
\hline  CNN-GRU \cite{sabir2019recurrent} & 97.6 & 47.6 & 85.8 & 86.6 & 79.4 \\
LipForensics-Scratch \cite{haliassos2021lips} & 93.0 & 56.7 & 98.8 & \underline{98.3} & 86.7 \\
LipForensics \cite{haliassos2021lips} & 99.7 & 90.1 & $\mathbf{99.7}$ & $\mathbf{99.1}$ & 97.1 \\
FTCN \cite{zheng2021exploring} & \underline{99.8} & \underline{99.6} & 98.2 & 95.6 & 98.3 \\
AltFreezing \cite{wang2023altfreezing} & \underline{99.8} & $\mathbf{99.7}$ & 98.6 & 96.2 & 98.6 \\
\hline UniForensics(ours) & $\mathbf{100.0}$ & 99.1 & \underline{99.0} & 97.0 & $\mathbf{98.8}$ \\
\hline
\end{tabular}
\label{tab:crossmanipulation}
\end{table}

\subsection{Generalization to Unseen Manipulation}

In real-world scenarios, it's often difficult to anticipate the methods used for face manipulation. So it is important for our model to have a strong generalization capability to unseen manipulations. We conduct experiments on FF++ \cite{rossler2019faceforensics++} with a leave-one-out setting, following previous works\cite{haliassos2021lips,wang2023altfreezing}. There are four types of manipulated face videos, i.e., DF, F2F, FS, and NT in FF++. We choose three of subsets as the training set, and use the remaining subset for evaluating the generalization capability of the model.

We compare the proposed method with state-of-the-art methods in Table \ref{tab:crossmanipulation}, and report the AUC(\%) scores. The AUC scores demonstrate that our model can achieve impressive performance on the whole FF++ test set (average AUC: 98.8\%), especially on the subsets DF (100.0\%) compared to previous methods. We also observed that LipForensics \cite{haliassos2021lips} achieves best performance on F2F and NT, but exhibits comparatively poor performance on FS. One possible explanation is that LipForensics employs a pre-trained model with a strong prior knowledge of the mouth region. When the manipulation quality of the lip region in the video is low, it can perform well; however, its performance is poor when the quality of mouth-region manipulation is high. Our method exhibits significantly greater stability across the four tampering techniques compared to LipForensics, due to our utilization of information from the entire facial region as the basis for assessment. FTCN and AltFreezing also perform excellently because they also emphasize the utilization of both temporal and spatial information. Our model achieves better performance over them in terms of the average AUC. This is mainly because our model benefits from the high-level semantic information of face encoder while their models are trained from scratch.

\begin{table*}
\normalsize
\caption{Cross-dataset evaluation. We report the AUROC ($\%$) on three unseen datasets: Celeb-DFv2, DFDC, FaceShifter. The models are trained on FaceForensics++ (FF++). The results of prior methods are directly cited from the original paper and their subsequences for fair comparison. \textbf{Bold} and \underline{underlined} values correspond to the best and the second-best value, respectively.}
\centering
\begin{tabular}{lcccccc}
\hline 
\multirow{2}*{Method} & \multirow{2}*{Venue} & \multirow{2}*{Input Type} & \multicolumn{4}{c}{Test Set AUC(\%)} \\
\cmidrule(lr){4-7} & & & Celeb-DFv2 & DFDC & FaceShifter & Avg \\
\hline 
Xception \cite{rossler2019faceforensics++} & ICCV 2019 & Frame & 73.7 & 70.9 & 72.0 & 72.2 \\
CNN-aug \cite{wang2020cnn} & CVPR 2020 & Frame & 75.6 & 72.1 & 65.7 & 71.1 \\
PatchForensics \cite{chai2020makes} & ECCV 2020 & Frame & 69.6 & 65.6 & 57.8 & 64.3 \\
Multi-task \cite{nguyen2019multi} & BTAS 2019 & Frame & 75.7 & 68.1 & 66.0 & 69.9 \\
Two-branch \cite{masi2020two} & ECCV 2020 & Frame & 76.7 & - & - & - \\
Face X-ray \cite{li2020face} & CVPR 2020 & Frame & 79.5 & 65.5 & 92.8 & 79.3 \\
SLADD \cite{chen2022self} & CVPR 2022 & Frame & 79.7 & - & - & - \\
SeeABLE \cite{larue2023seeable} & CVPR 2023 & Frame & 87.3 & 75.9 & - & - \\
SBI-R50 \cite{shiohara2022detecting} & CVPR 2022 & Frame & 85.7 & - & 78.2 & - \\
EFNB4 + SBIs \cite{shiohara2022detecting} & CVPR 2022 & Frame & \underline{93.2} & 72.4 & 98.3 & 88.0 \\
AUNet \cite{bai2023aunet} & CVPR 2023 & Frame & 92.8 & 73.8 & - & - \\
\hline 
CNN-GRU \cite{sabir2019recurrent} & CVPRW 2019 & Video & 69.8 & 68.9 & 80.8 & 73.2 \\
LipForensics-Scratch \cite{haliassos2021lips} & CVPR 2021 & Video & 62.5 & 65.5 & 84.7 & 70.9 \\
LipForensics \cite{haliassos2021lips} & CVPR 2021 & Video & 82.4 & 73.5 & 97.1 & 84.3 \\
RealForensics \cite{haliassos2022leveraging} & CVPR 2022 & Video & 86.9 & 75.9 & $\mathbf{99.7}$ & 87.5 \\
FTCN \cite{zheng2021exploring} & ICCV 2021 & Video & 86.9 & 74.0 & 98.8 & 86.6 \\
LTTD \cite{guan2022delving} & NeurIPS 2022 & Video & 89.3 & $\mathbf{80.4}$ & \underline{99.5} & \underline{89.7} \\
ISTVT \cite{zhao2023istvt} & TIFS 2023 & Video & 84.1 & 74.2 & 99.3 & 85.9 \\
AltFreezing \cite{wang2023altfreezing} & CVPR 2023 & Video & 89.5 & - & 99.4 & - \\
TALL \cite{xu2023tall} & ICCV 2023 & Video & 90.8 & 76.8 & $\mathbf{99.7}$ & 89.1 \\
\hline UniForensics & ours & Video & $\mathbf{95.3}$ & \underline{77.2} & 99.1 & $\mathbf{90.5}$ \\
\hline
\end{tabular}
\label{tab:crossdataset}
\end{table*}

\subsection{Generalization to Unseen Datasets}

In practice, we are confronted not only with unknown forgery methods but also with diverse and complex video quality and shooting environments. These always lead to significant performance degradation of known methods. Therefore, it is critical to evaluate the generalization capability of models to unseen datasets. To evaluate the generalization ability of our model, we use the original videos and all four types of fake videos in FF++\cite{rossler2019faceforensics++} as the training data, then evaluate the performance on 3 challenging datasets, namely Celeb-DFv2\cite{li2020celeb}, DFDC\cite{dolhansky2020deepfake} and FaceShifter\cite{li2020advancing}. 

We report the AUC results in Table \ref{tab:crossdataset}. 
To ensure objectivity, we selected the best models at both image and video levels for comparison. We observe that our method achieves the best performance on CDF (95.3\%) and average AUC of the three datasets (90.5\%), second best performance on DFDC (77.2\%) and competitive performance on FSh (99.1\%). It is worth noting that all other methods perform unsatisfactorily on CDF, while our model outperforms the state-of-the-art methods on CDF by 2.1\%, which shows its strong generalization capability. We observed that most image-level methods exhibit poor performance on the FSh while several video-level methods perform well. It might result from that temporal information provides significant assistance in detecting forgery methods similar to FaceShifter, thus proving the superiority of spatio-temporal joint detection.

\subsection{Robustness to Video Corruptions}

Video files are often affected by video corruptions such as compression and noise during their dissemination on the internet. Therefore, is is significant for deepfake detectors to possess robustness against common video corruptions. We investigate the robustness of our models by training on uncompressed FF++ and then testing on FF++ samples that were applied with various video corruptions. We consider the following operations at five severity levels, as given in \cite{jiang2020deeperforensics}: changes in saturation, changes in contrast, adding block-wise distortions, adding White Gaussian noise, blurring, pixelation, and applying video compression (H.264 codec).

\begin{figure*}[t]
  \centering
  \includegraphics[width=1.0\linewidth]{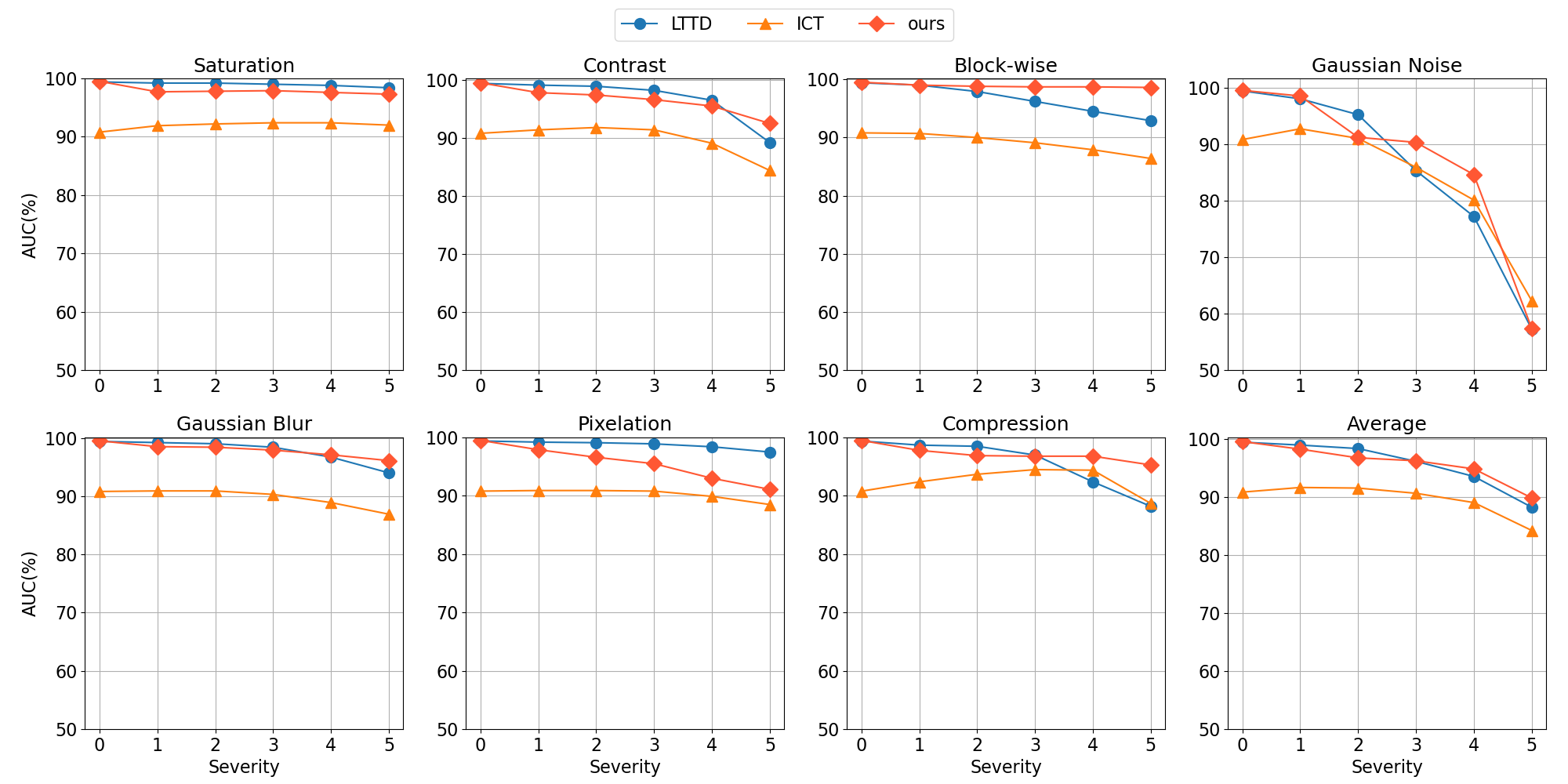}

   \caption{Robustness evaluation. We report the AUC (\%) scores of our methods under five different levels of seven particular types of corruption. ”Average” denotes the mean across all corruptions at each severity level.}
   \label{fig:ptb}
\end{figure*}

In Fig. \ref{fig:ptb}, we show results of increasing the severity of each corruption. Comparing with other methods, our approach shows excellent robustness against most of the perturbations. We also present the comparison with four state of the arts in Table \ref{tab:robustness}, where the results are averaged from all five levels. Our model is overall more robust to those visual perturbations on average compared with these methods. This is attributed to the introduction of pretrained face encoder, which can provide high-level semantic features and mitigate overfitting to low-level artifacts.

\begin{table*}
\normalsize
\caption{Robustness evaluation. Average performance evaluated on perturbed videos at five levels. Clean: origin videos, CS: color saturation, CC: color contrast, BW: block-wise noise, GNC: gaussian noise, GB: gaussian blur, PX: pixelation, VC: video compression, Avg: averaged performance on distorted videos. We report the AUC(\%) scores.}
\centering
\begin{tabular}{lccccccccc}
\hline Method & Clean & CS & CC & BW & GNC & GB & PX & VC & Avg \\
\hline Face X-ray \cite{li2020face} & 99.8 & 97.6 & 88.5 & $\mathbf{99.1}$ & 49.8 & 63.8 & 88.6 & 55.2 & 77.5 \\
LipForensics \cite{haliassos2021lips} & $\mathbf{99.9}$ & $\mathbf{99.9}$ & $\mathbf{99.6}$ & 87.4 & 73.8 & 96.1 & 95.6 & 95.6 & 92.6 \\
ICT \cite{dong2022protecting} & 90.8 & 92.2 & 89.6 & 88.8 & 82.4 & 89.6 & 90.2 & 92.7 & 89.4 \\
LTTD \cite{guan2022delving} & 99.4 & 98.9 & 96.4 & 96.1 & 82.6 & 97.5 & $\mathbf{98.6}$ & 95.0 & 95.0 \\
\hline UniForensics(ours) & 99.5 & 97.7 & 96.0 & 98.8 & $\mathbf{84.4}$ & $\mathbf{97.6}$ & 94.8 & $\mathbf{96.7}$ & $\mathbf{95.1}$ \\
\hline
\end{tabular}
\label{tab:robustness}
\end{table*}

\subsection{Ablation Study}

\subsubsection{Effect of Different Pretrained Models} To investigate the effectiveness of the proposed model initialization strategy, we compared with our settings using the following different pre-trained models: (1) ViT-based face encoder of FaRL, which is an image-level model. It contains abundant prior knowledge of faces that can benefit detection task. (2) UniV2 with CLIP pretrained ViT, which differs from ours in that it employs pretraining on general image processing tasks. (3) The official released UniV2 model initialized with CLIP and finetuned on K710 video classification dataset. For fair comparison, we do not use any synthetic samples and train all the models in a single-stage scheme. The results are shown in in Table \ref{tab:pretrain models}. 

The variants we compared stand for several pretrain scheme: (1) Simply finetune a pretrained face encoder on image level as deepfake detector. (2) Initialize a spatio-temporal model with a pretrained general image encoder and finetune on video level. (3) Finetune a general video classifier trained on general video classification tasks. From Table \ref{tab:pretrain models}, we show that our scheme surpasses the others by 3.3\% in terms of average AUC. The variant of finetuned FaRL performs the second best, due to its strong prior knowledge of faces. Our method performs much better than it mainly because we can leverage temporal information, which image-level models cannot achieve. It can also be observed that our method outperforms CLIP-initialized model, indicating that pretrained face encoder is superior to general image encoder as part of a deepfake detector. The model initialized with CLIP and finetuned on K710 does not perform well, even worse than the one without finetuning. This can be attributed to that K710 is a video classification task rather than a video encoding task. Consequently, the model learns features that are not sufficiently generalizable and contains excessive bias from K710 classication.

\begin{table}
\caption{Effect of different pretrained model. FaRL is a face encoder. CLIP+UniV2 is the UniV2 model initialized with CLIP. UniV2 (K710) is initialized with CLIP and finetuned on K710.}
\centering
\begin{tabular}{lcccc}
\hline Pretrained Model & Celeb-DFv2 & DFDC & FaceShifter & Avg \\
\hline FaRL (image) & 86.3 & 67.9 & $\mathbf{98.6}$ & 84.3 \\
CLIP+UniV2 & 82.8 & 66.2 & 95.6 & 82.1 \\
UniV2 (K710) & 79.0 & 69.2 & 95.6 & 81.5 \\
FaRL+UniV2 (ours) & $\mathbf{91.5}$ & $\mathbf{72.8}$ & 98.4 & $\mathbf{87.6}$ \\
\hline
\end{tabular}
\label{tab:pretrain models}
\end{table}

\subsubsection{Effect of Dynamic Video Self-blending} We compare our dynamic video self-blending with the two naive strategies mentioned earlier. (1) Static: apply same transformation $K^{(t)}$ to every frame, making the forgery characteristics static over time. (2) Independent: randomly sample the transformation $K^{(t)}$ for each frame, making them independent of each other. For the sake of simplicity and clarity, we use single-stage training scheme for the three. We also compared our method with the scenario where no generated negative samples were used. Furthermore, we control the frequency of dynamic video self-blending by changing the sample interval of  Gaussian filter's mean. We choose three variants as follows: high frequency where $\mu\sim U[3,9]$, intermediate-frequency where $\mu\sim U[0,6]$ and low frequency where $\mu\sim U[-3,3]$. 

As shown in Table \ref{tab:dvsb ablation}, it is obvious that synthetic samples of each variants will lead to performance improvements. As expected, our dynamic video self-blending surpasses two naive approaches in terms of average AUC score, 
which demonstrates that the negative sample generated with DVSB exhibits more manifold features in the temporal domain. We can also observe that DVSB achieves the best performance with the low frequency setting. One possible explanation is that $\mu\sim U[-3,3]$ is closer to the generation process of forged samples in real world.

\begin{table}
\caption{Effect of dynamic video self-blending. Compared with two naive strategies: static and independent, no synthetic training and three variant of DVSB with different frequency setting.}
\centering
\begin{tabular}{lcccc}
\hline Temporal Manipulation & CDF & DFDC & FSh & Avg \\
\hline no synthetic & 87.4 & 71.5 & 98.2 & 85.7 \\
static & 91.5 & $\mathbf{72.8}$ & 98.4 & 87.6 \\
independent & 90.8 & 71.0 & 98.7 & 86.8 \\
dynamic (high frequency) & 91.2 & 70.4 & 98.5  & 86.7 \\
dynamic (intermediate-frequency) & 91.7 & 71.9 & $\mathbf{99.1}$ & 87.6 \\
dynamic (ours, low frequency) & $\mathbf{93.3}$ & 72.5 & $\mathbf{99.1}$ & $\mathbf{88.3}$ \\
\hline
\end{tabular}
\label{tab:dvsb ablation}
\end{table}

We also conduct ablation studies on the MAdaIN used in our DVSB. The results are reported in Table \ref{tab:madain}. It indicates that incorporating MAdaIN results in a 1.4\% performance gain in terms of average AUC score. This demonstrates that the addition of MAdaIN indeed can generate more realistic forged samples, thereby increasing the learning difficulty for the model and guiding it towards learning features with greater generalizability.

\subsubsection{Effect of Two-stage Training Scheme} We conduct ablation study to analyze the effect of our two-stage training scheme. For fairness, we compare our result with single-stage training using FF++ training set and real faces in Vox2 for synthesizing fake samples. We also investigated the impact of incorporating additional genuine human faces of Vox2 under the single-stage strategy. The result reported in Table \ref{tab:two stage} demonstrate that our two-stage training scheme outperforms the single-stage scheme on CDF, DFDC and average AUC by 1.6\%, 2.7\% and 1.5\% points, using the same data of FF++ and Vox2. Surprisingly, the simple addition of extra training data in the single-stage strategy did not lead to a significant improvement in performance. This suggests that the improvement in performance primarily stems from the introduction of our forgery process identification in the first training stage.

We also conducted ablation studies on the sample strategy of positive pairs in the self-supervised pretraining stage. From Table \ref{tab:two stage}, it can be observed that using positive pairs sampled from the whole parameter space of $\{K_c, K_q, K_d, K_g, K_a, K_b\}$ leads to a sharp decline in performance, even worse than the single-stage training without data from Vox2. We argue that this is because when we designate samples with the same $K_c$ and $K_q$ as positive pairs, the model will regard the differences between target and source image as forgery-specific features. Nevertheless, the differences in color distribution and image quality between the source and target are primarily attributed to the inherent characteristics of the images. Even in actual forgery processes, the same forgery method applied to different target-source image pairs yields different $K_c$ and $K_q$ values. So this will only increase the difficulty of learning for the model and may even result in the model's learning incorrect features. It is noteworthy that in our experiments, we found that it prevents the model from converging in the first training stage. Conversely, if we sample from $\{K_d, K_g, K_a, K_b\}$ to generate positive sample pairs, while allowing parameters $K_c$ and $K_q$ to vary randomly, the model learned rich and highly generalizable features, making a 7.3\% increase in performance.

\begin{table}
\caption{Effect of using MAdaIN in DVSB. We compare the performance with and without MAdaIN. The best results are in \textbf{bold}.}
\centering
\begin{tabular}{lcccc}
\hline Pretrained Model & Celeb-DFv2 & DFDC & FaceShifter & Avg \\
\hline DVSB w/o MAdaIN & 94.2 & 73.9 & $\mathbf{99.2}$ & 89.1 \\
DVSB w/ MAdaIN (ours) & $\mathbf{95.3}$ & $\mathbf{77.2}$ & 99.1 & $\mathbf{90.5}$ \\
\hline
\end{tabular}
\label{tab:madain}
\end{table}

\begin{table}
\caption{Effect of two-stage training scheme. Compared with single-stage training with and without data from Vox2. * indicates sampling from the whole parameter space of $K$.}
\centering
\begin{tabular}{lcccc}
\hline Training Scheme & Celeb-DFv2 & DFDC & FaceShifter & Avg \\
\hline single-stage w/o Vox2 & 93.3 & 72.5 & $\mathbf{99.1}$ & 88.3 \\
single-stage w/ Vox2& 93.7 & 74.5 & 98.9 & 89.0 \\
two-stage* & 86.5 & 66.6 & 96.4 & 83.2 \\
two-stage (ours) & $\mathbf{95.3}$ & $\mathbf{77.2}$ & $\mathbf{99.1}$ & $\mathbf{90.5}$ \\
\hline
\end{tabular}
\label{tab:two stage}
\end{table}

\subsubsection{Design of Model Architecture} In the local blocks of our model, the inclusion of the 3D convolution module played a significant role, as it introduced temporal information comparison in the calculation of attention in the spatial domain. Without the use of 3D convolution, the model would only perform attention calculation in the spatial domain, which is similar to a ViT model. We conducted experiments to further substantiate the significance of integrating 3D convolution in local blocks. As shown in Table \ref{tab:local block}, we experimented with solely utilizing spatial attention calculation, applying 3D convolution before attention, applying 3D convolution after attention, and adding 3D convolution both before and after attention modules. 

The results reported demonstrate that adding 3D convolution both before and after attention modules outperforms other variants. And the absence of the 3D convolution module would result in an overall decline in the model's detection performance, which aligns with our expectations. If we consider solely from the perspective of the pathway composed of concatenated local blocks, the two variants which pair only one 3D convolution with every attention module are similar. They both insert a 3D convolution between each attention module of the pretrained face encoder. Their distinction lies only in the first and last local blocks. This should be the primary reason for their similar performance outcomes. Similarly, the approach utilizing dual 3D convolutions can be considered as inserting two 3D convolutions between each attention module of the pretrained face encoder. The superior performance of it is primarily due to its largest parameter count.

\begin{figure*}[t]
  \centering
  \includegraphics[width=1.0\linewidth]{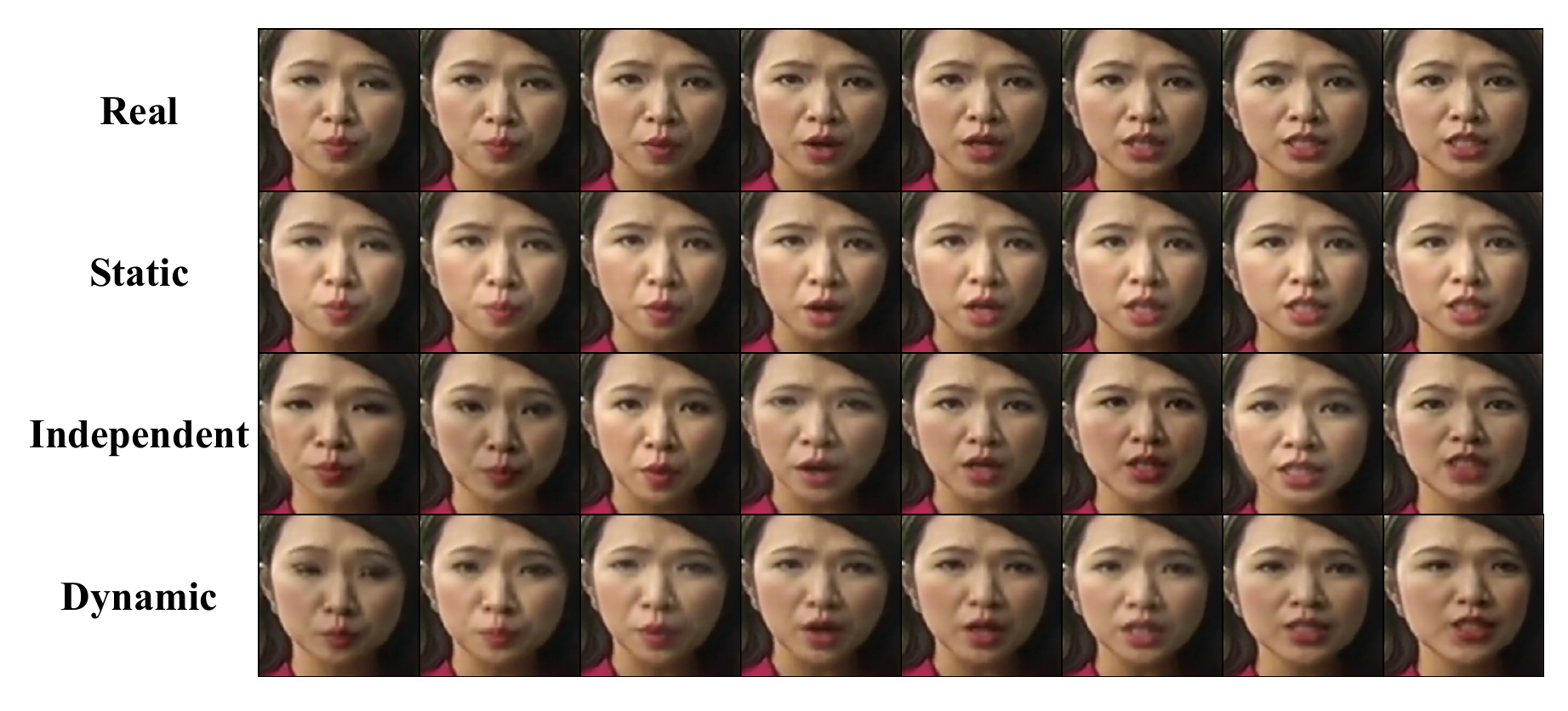}

   \caption{Examples of different temporal manipulations on the same video clip. Real: real clip without manipulations. Static: apply same transformation to each frame in one clip, Independent: randomly apply independent transformation to each frame, Dynamic: the proposed dynamic video self-blending(DVSB).}
   \label{fig:dvsb}
\end{figure*}

\begin{table}
\caption{Effect of inserting 3D convolution in the local blocks. Variants involving different positions and numbers of 3D convolution insertions were compared.}
\centering
\begin{tabular}{lcccc}
\hline Variants & Celeb-DFv2 & DFDC & FaceShifter & Avg \\
\hline attn only & 92.7 & 71.1 & 98.9 & 87.6 \\
conv3D+attn & 92.9 & 73.4 & 98.3 & 88.2 \\
attn+conv3D & 92.8 & $\mathbf{73.5}$ & 98.3 & 88.2 \\
conv3D+attn+conv3D & $\mathbf{93.3}$ & 72.5 & $\mathbf{99.1}$ & $\mathbf{88.3}$ \\
\hline
\end{tabular}
\label{tab:local block}
\end{table}

We also conduct ablation studies on the positions and numbers of global blocks used in the model. The role of the global blocks is to integrate and compare global spatio-temporal information, thus capturing the inconsistency in both the temporal and spatial domains. Due to the fixed structure of the pretrained ViT model we used, the number of local blocks in our model is fixed at 12. For each local block, we have the option to either add a corresponding global block or not. We conducted comparative experiments on the following variants of structure: (1) adding corresponding global blocks to the 3rd, 6th, 9th, and 12th local blocks, totaling 4 global blocks; (2) adding corresponding global blocks to the 7th, 8th, 9th, 10th, 11th, and 12th local blocks, totaling 6 global blocks; (3) adding corresponding global blocks to every local block, totaling 12 global blocks; (4) adding corresponding global blocks to the 2th, 4th, 6th, 8th, 10th, and 12th local blocks, totaling 6 global blocks.

We reported the results in Table \ref{tab:global block}. It can be observed that utilizing a smaller number of global blocks, such as 4, results in performance that is somewhat inferior to using 6 or more. By comparing settings (2) and (4), we can ascertain that, with the same number of global blocks, an even distribution corresponding to both shallow and deep local blocks yields better results than corresponding solely to deep layers. One possible explanation is that facial semantic features from different levels helps the model distinguish authenticity and fake from multiple perspectives. When the number of global blocks reaches 6 or more, the performance of the model tends to saturate, as can be deduced from the results of the third and fourth rows in Table \ref{tab:global block}. Therefore, we believe that setting the number of global blocks to 6 can achieve a better balance between performance and computational cost. Therefore, in other experiments, we consistently adopt setting (4).

\begin{table}
\caption{Effect of different positions and
numbers of global blocks used in the model.}
\centering
\begin{tabular}{lcccc}
\hline Global Blocks & Celeb-DFv2 & DFDC & FaceShifter & Avg \\
\hline $\{3,6,9,12\}$ & 91.8 & 72.0 & 98.5 & 87.5 \\
$\{7,8,9,10,11,12\}$ & 92.0 & $\mathbf{73.6}$ & 98.8 & 88.1 \\
$\{1,2,...,12\}$ & $\mathbf{93.4}$ & 72.6 & 98.0 & $\mathbf{88.3}$ \\
$\{2,4,6,8,10,12\}$ & 93.3 & 72.5 & $\mathbf{99.1}$ & $\mathbf{88.3}$ \\
\hline
\end{tabular}
\label{tab:global block}
\end{table}

\subsubsection{Visualization}To visually demonstrate the proposed dynamic video self-blending, we present examples of different temporal manipulations. As shown in Fig. \ref{fig:dvsb}, we presented authentic video clips, counterfeit samples generated by the two naive strategies mentioned earlier, and fake samples produced using our dynamic video self-blending(DVSB) method. From the visualization results, we can observe that the static strategy, which applies same transformation $K^{(t)}$ to every frame, exhibits a lack of temporal variation. Conversely, if we look at the independent strategy, which randomly samples the transformation $K^{(t)}$ for each frame, there is excessively frequent and significant jitter in the clip. Both of them produce overly monotonous temporal features, thus making them easy to detect. Compared to them, it can observed that the samples generated by DVSB maintain relatively stable forgery features while embodying variations in temporal sequence. Therefore, samples generated using our DVSB exhibit richer and more realistic forgery features especially in temporal domain, thereby enhancing the model's detection performance.

\begin{figure}[t]
  \centering
  \includegraphics[width=0.9\linewidth]{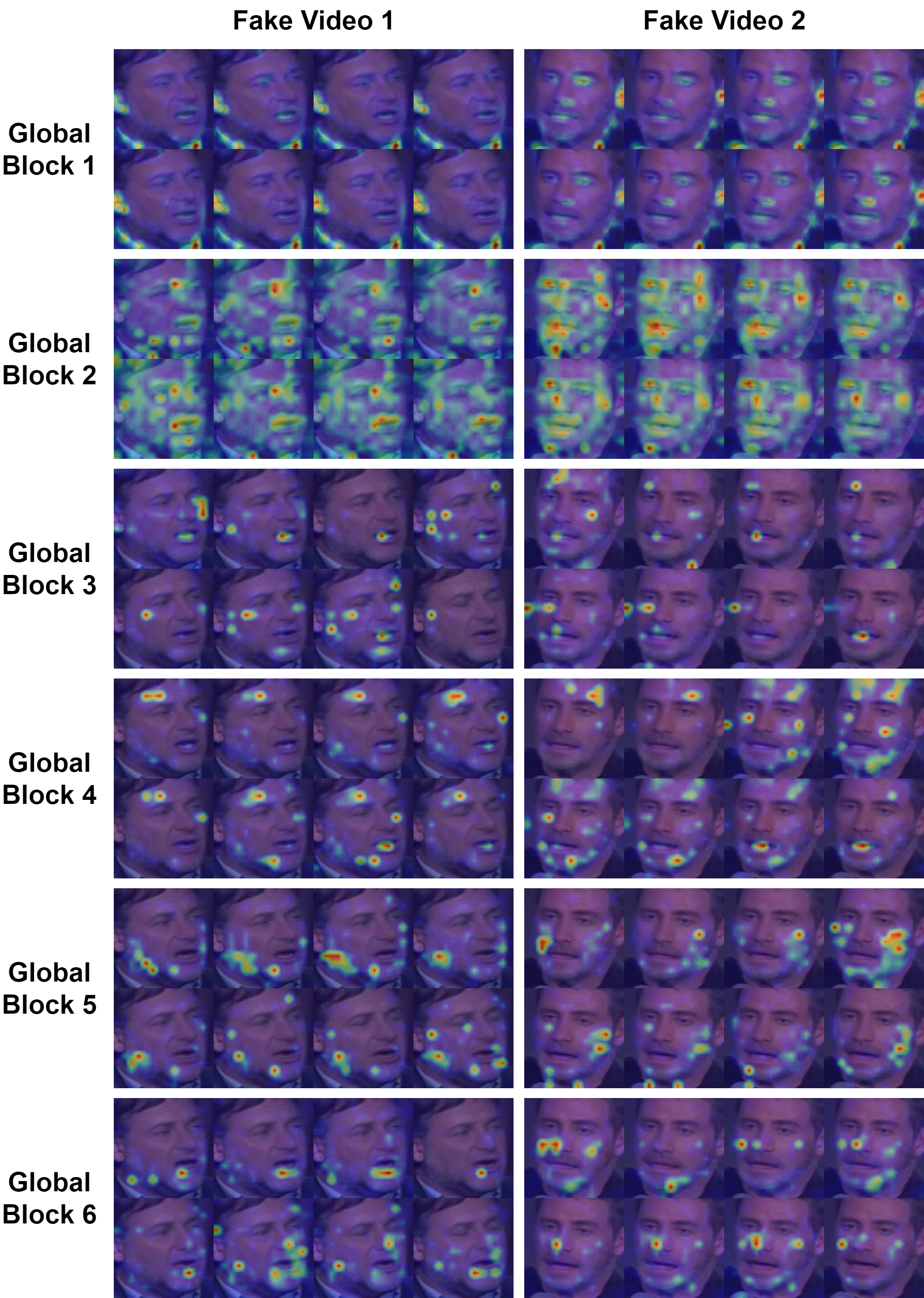}

   \caption{Attention maps of global blocks. We choose two fake video clips and show the attention maps of global block 1 to 6.}
   \label{fig:attn}
\end{figure}

We also visualized the global attention within the global blocks. As shown in Fig. \ref{fig:attn}, we choose two fake videos and show their attention map from global block 1 to 6. It can be observed that the global blocks at each layer focus on different regions while simultaneously performing comparisons over time. Each global block receives features from its corresponding local block, which includes a certain layer of pretrained face encoder. Therefore, each global block discriminates based on different levels of facial semantic information and focuses on different region of the frames. This visualization shows the model's ability to compare temporal information. Also, it further identifies that facial semantic features at various levels assist the model in discerning the truth from falsehood from multiple perspectives.

\section{Conclusion}
In this paper, we propose a new face forgery detection framework called UniForensics, which leverages a transformer-based video classification network and is initialized with a meta-functional face encoder for enriched facial representation. Furthermore, we propose dynamic video self-  blending (DVSB) to enrich the temporal forgery artifacts of synthesized fake videos, thus preventing overfitting. Based on DVSB, we also design a two-stage training approach: The first stage employs a novel self-supervised contrastive learning, the network is encouraged to focus on forgery traces by impelling videos generated by the same forgery process to have similar representations. The second stage involves fine-tuning on face forgery detection dataset
to build a deepfake detector on the basis of the representation learned in the first stage. Compared with the previous SOTA deepfake detection methods, UniForensics significantly improves the performance and robustness. The proposed two-stage training strategy and our dynamic video self-blending do not impose specific requirements on the model structure. In future work, we will further explore the effectiveness of our method across different spatio-temporal models.

\bibliographystyle{IEEEtran}
\bibliography{IEEEabrv,main}

\end{document}